\documentclass[oribibl]{llncs}

\usepackage{graphicx}
\usepackage{caption}

\begin{document}

\title{Robot Dream}

%

\author{
Alexander Tchitchigin\inst{1}
\and  Max Talanov\inst{2}
\and Larisa Safina\inst{3}
\and Manuel Mazzara\inst{4}
}

\institute{
    Innopolis University, Russia,\\
    \email{a.chichigin@innopolis.ru}\\
    Kazan Federal University, Russia,\\
    \email{a.tchichigin@it.kfu.ru}\\
\and
    Kazan Federal University, Russia,\\
    \email{max.talanov@gmail.com}
\and
   Innopolis University, Russia,\\
   \email{l.safina@innopolis.ru}
\and
    Innopolis University, Russia,\\
    \email{m.mazzara@innopolis.ru}
}

\maketitle

\begin{abstract}

In this position paper we present a novel approach to neurobiologically plausible
implementation of emotional reactions and behaviors for real-time autonomous robotic systems.
The working metaphor we use is the ``day'' and ``night'' phases of mammalian life.
During the ``day'' phase a robotic system stores the inbound information and is controlled by
the light-weight rule-based system in real time. In contrast to that, during the ``night'' phase the
stored information is been transferred to the supercomputing system to update the realistic
neural network: emotional and behavioral strategies.


\end{abstract}

\section{Introduction}\label{intro}

Some time ago we have asked ourselves,
``Why could emotional phenomena be so important for robots and AI (artificial intelligence) systems?''
As for humans: emotional mechanisms manage processes like attention, resource allocation, goal
setting, etc.
These mechanisms seems to be beneficial for computational systems in general and therefore for
AI and robotic systems.
Still these phenomena tend to be ``difficult'' for computational as well as AI and robotics
researchers. To implement all phenomena related to emotions we have decided to use
simulation of nerobiological processes.

Besides, without understanding and simulation of emotions the effective AI--human
communication is almost impossible.
Let's review one real life example.
Imagine a concierge robot in a hotel, finishing checking in a
fresh lodger, when a man from room 317 rushes into concierge desk in a
panic. ``Water tap is broken and it starts flooding the whole floor!''
Even in a case when it is optimal to finish check-in first,
it is very likely that the new guest would
wish a concierge to switch immediately to arisen problem. That happens
due to empathy between human beings. If a robot does not understand
emotions, it can not prioritize tasks appropriately, which would damage
client's satisfaction and hotel's reputation.

We could identify the main contribution of this paper as an approach that allows
to expand practical autonomous real-time control of a robotic platform with realistic
emotional appraisal and behavior based on spiking neural network simulation and neuromodulation.

In section \ref{the-problem} we substantiate the need for neurobiologically plausible
emotional simulation and point out the mismatch between computational resources
available to current robotic systems and what is required for neuronal simulation.
In section \ref{my-idea} we introduce our concept how a robotic system execution
can be separated into ``day'' and ``night'' phases in order to bridge the gap
between a robotic system and supercomputer performing the simulation.
In section \ref{the-details} we introduce the notion of ``bisimulation'' to answer
the questions of learning and mapping from realistic neural network to rules-based
control system.
Section \ref{related-work} provides the information about the actual topics
in the field of affecting computations, notable authors and research projects
in this area.
Finally we sum up the ideas presented in the paper and discuss the arose questions
with attention to the steps we are going to take in order to resolve them in
section \ref{conclusion}.

\section{The Problem}\label{the-problem}

There are several cognitive architectures that implement emotion phenomena,
some of the most notable are listed in section \ref{related-work}.
Rather than implementing the emotional model in a computational system, we re-implemented the
neurobiological basis using simulation.
This was done to create a biologically plausible approach and to validate the results of
our simulations from neurobiological perspective.
The other way around could not provide proper evidence that the result could be
regarded as emotional phenomenon.
We used the model of basic mechanisms of a mammalian brain via neuromodulation
and their mapping to basic affective states \cite{lovheim2012,tomkins1962,tomkins1962_2,tomkins1963,tomkins1981}.
We used the realistic spiking neural networks with neuromodulation reconstructing
all brain structures involved into the pathways of neuromodulators of the ``cube of emotions'' by Hugo L\"{o}vheim \cite{lovheim2012}.
Unfortunately, current robotic systems usually do not have enough memory and computational capacity
to run realistic simulations of human brain activity.

For example, this is computational resources of rather advanced bipedal
robotic platform AR-601:

\begin{itemize}
\item CPU --- 4th Gen Intel® Core™ i7-4700EQ 4-Core 3.4GHz processor;
\item System Memory --- 1 x204-Pin DDR3L 1333MHz SO-DIMM up to 8 GB;
\end{itemize}

However the simulation of 1\% of human brain required a cluster of 250
K-supercomputers (each contains 96 computing nodes, each node contains a
2.0 GHz 8-core SPARC64 processor and 16 GB of memory) that was done by
RIKEN institute in 2013 and this simulation was slower than human brain
in 1000 times \cite{RIKEN}. According to the estimates
of the Human brain project the computational capacity to simulate whole human brain should be 30 exaflop that is not feasible at the moment.

\section{Day phase and Night phase}\label{my-idea}

To enable (soft) real-time behaviour from a robotic system it should run
some kind of traditional rule-based control system. We do not constrain
type or architecture of this system, it might be some sort of Boolean
expert system, fuzzy-logic based or Bayesian-logic based system,
possibly augmented with Deep-learning pre-processing cascades for visual
and/or audio input channels. In our approach robot's control system
should not only produce appropriate reaction for input stimulus but also
record these stimuli for post-processing during ``night phase''.

So we propose that robotic system records and stores all input signals for further
post-processing. It might store either raw inputs (video and audio
streams for example) or some higher-level aggregated form like
Deep-learning cascade output (if control system uses such pre-processing
mechanisms). And when robot ``goes to sleep'', it transmits all new
recorded inputs to the supercomputer over some network connection.
Actually robotic system might not ``sleep all night long'', it can stay
active while supercomputer processes new information, we assume only
that the robot regularly ``takes a break'' to transmit recorded
information and later to receive results and update rules of its
control system.

\begin{figure}[ht]
  \centering
  \includegraphics[width=0.9\textwidth]{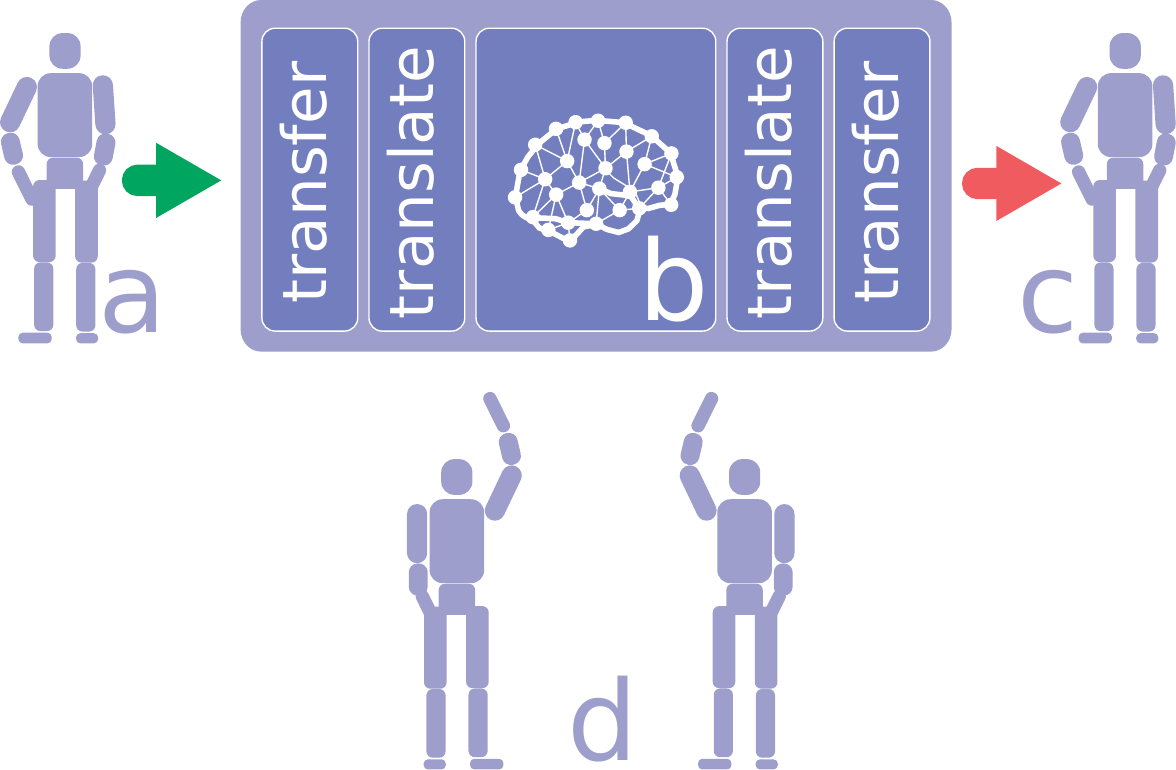}
  \captionsetup{singlelinecheck=off}
  \caption[foo]{ Night phase and Day phase.
  \begin{description}
  \item[a.] In this position a robotic system transfers the accumulated experience into spiking neural network
  \item[b.] Processing is done as follows:
      \begin{enumerate}
      \item First the accumulated experience is transferred from a robotic system to the processing center
      \item Then bisimulation starts producing a set of updated rules for robot's control system
      \item Finally update is transferred to the robotic system
      \end{enumerate}
  \item[c.] The updated rules of the control system are transferred to the robotic system and applied to it
  \item[d.] The robotic system runs updated learning, appraisal and behavioral strategies in current environment and accumulate new experience to be processed again starting from a
  \end{description}
  }
  \label{fig:robot-dream}
\end{figure}

Finally, the supercomputer runs realistic spiking neural network that
simulates brain regions involved into neuromodulation pathways. This
network receives as an input actual signals recorded by robotic system
in a raw or pre-processed (aggregated) form. Then the network processes
inputs in the simulated time (which might be faster or slower than clock
time depending on available computational power), so we can infer emotional
response of the system \cite{talanov2014, bica2015neucogar} from resulting
neural activity and neuromodulators levels. From emotional responses of spiking neural
network we generate updates to the rules of a control system which are
sent back to the robot.

\section{Bisimulation}\label{the-details}

With the approach of simulating separately a reaction of realistic
neural network to input stimuli in order to infer an appropriate emotional
response, two essential questions arise. First of all, how are we going to
train our spiking neural network so it produces relevant and suitable
emotional response, and after that how are we going to understand what
particular rules of robot's control system we need to update to
accommodate emotional mismatch if any?

To answer both questions we propose an approach that we call
bisimulation by analogy with the notion in state transition systems. The
idea is to simultaneously run both realistic neural network and a copy
of robot's control system on the same input. This immediately gives an
answer to the second question as we can see which rules of the control
system gets activated on current input and what emotional response they
imply so we can compare that with emotional reaction inferred from
spiking neural network and make necessary adjustments.

But we can use bisimulation to address the question of spiking neural
network training too. If we have a rule-based control system with
reasonably good emotional reactions we can use it as a source of reward
stimuli for neural network. So initially we start running both neural
network and control system on sample input and whenever control system
produce positive emotional value of the current stimulus, we increase
dopamine level in neural network simulating positive feedback
(analogously to reward reaction (e.g.~to feeding) in mammals).
Respectively when control system signals negative emotional value, we
decrease dopamine level in neural network.

After realistic neural network learns to mimic emotional responses of
control system it can evolve emotional reactions in unsupervised
way as it happens in mammals. From that point we can start adjusting
emotional reactions of the control system to ever improving ones of
spiking neural network.

Speaking of adjusting the rules of a control system with respect to emotional
feedback from neural network we highlight several aspects of decision-making that
undergo influence of emotional state \cite{roleofemotions, talanov2015}:

\begin{itemize}
 \item decisive / less decisive
 \item speed of decision-making
 \item bias to positive or negative ``thoughts''
 \item optimistic / conservative
 \item careful / risky
\end{itemize}

Let's discuss how these traits could affect a rules-based control system taking
as an example NARS framework \cite{wang2013nars}, based on non-axiomatic logic.
NARS system was specifically designed to work under bounded amount of knowledge,
computational resources and available memory. Therefore the system dynamically
redistributes resources among tasks and processes that are needed to be performed
according to assigned priorities and relevance. Hence NARS can spend less resources
on one task and get approximate answer and much more resources on another task, which leads to
obtaining a more precise solution.

Thus the most straightforward aspect to map is the speed of decision-making:
the more computational and memory resources we allocate to the whole NARS system
the faster it provides decisions to perform necessary actions and vice versa.
This could be implemented as some kind of CPU-boost in case of excitement or fear emotional states. Furthermore we can say that NARS have built-in
``conservativeness handle'': parameter $k$ in denominator of confidence formula.
The higher the value of $k$, the more unconfident the system about its knowledge,
the more effort it will spend to ascertain a solution. We can model inclination
of the system to ``positive'' or ``negative'' ``thoughts'' adjusting relative
priorities of facts with and without logical negation. Also risky behavior could
be modeled by lowering the confidence threshold of accepted actions.

Similar mapping for decision-making traits can be constructed for probabilistic
or fuzzy-logic based control systems, though in each particular case we have to
carefully consider available parameters and adjust the mapping according to
semantics and pragmatics of the control system we work with.

\section{Related work}\label{related-work}

Starting from the late 90-es the interest to emotions and emotional representations
in computational systems has been growing \cite{kismet,affectivecomputing,affectivecomputingchallanges,whatdoesitmeanforcomputer}.
This rise of activity is based on understanding of the role of emotions in human
intelligence and consciousness that was indicated by several neuroscientists
\cite{Damasio1998,Damasio1999,f5}. Starting from the seminal book by Rosalind Picard,
\cite{picard1997} (though we could mention earlier attempts to implement emotions in the machines \cite{breazeal2002}), we could identify two main directions in a new research field of affective computing: emotion recognition and re-implementation of emotions in a computational system. There are several cognitive architectures that are capable of the re-implementation of emotional phenomena in different extent, starting from SOAR \cite{laird2008} and ACT-R \cite{harrison2002} to modern BICA \cite{samsonovich2013}.
The interest in implementation of emotional mechanisms is based on the role
of emotional coloring in appraisal, decision making mechanisms, and emotional behavior,
indicated by Damasio in \cite{damasio1994}. Our approach takes a step
further on the road for neurobiologically plausible model of emotions \cite{bica2015neucogar}.

We have to mention other perspectives on emotional re-implementations in a computational machines:
Arbib and Fellous \cite{neuromodulatory, fellous2004} created the neurobiological
background for the direction to neurobiologically inspired cognitive architectures;
the appraisal aspects are in focus of Marsella and Gratch research
\cite{marsella2010, marsella2003, gratch2005} as well as in Lowe and Ziemke works
\cite{on_role_of_emotion, roleOfReinforcement}.
As the neuropsychological basement for our cognitive architecture we used ``Cube of emotions'' created by Hugo L\"{o}vheim \cite{lovheim2012}. It bridges the psychological and neurobiological phenomena in one relatively easy to implement programmatically model based on three-dimensional space of three neuromodulators: noradrenaline, serotonin, dopamine, which is its main advantage
from our perspective. For the implementation we used realistic neural network simulator NEST
\cite{Gewaltig:NEST} recreating the neural structures of the mammalian brain. As it was
mentioned earlier in this paper, the processing of the simulation took 4 hours of supercomputer's processing time to calculate 1000 milliseconds.

\section{Conclusion}\label{conclusion}

In our article we described an idea or approach for augmentation of
autonomous robotic systems with mechanisms of emotional revision and
feedback. There are many open questions: input formats for realistic
neural network, emotional revision thresholds and emotional equalizing
(homeostasis) and so on. On the one hand, different answers to these
questions allow adaptation of the approach to a range of possible
architectures of robots' control systems. On the other hand, even better
solution would be a software framework implementing our approach with several
pluggable adapters to accommodate the most popular choices for robots'
``brains''.

So our next step is to implement a proof-of-concept of proposed
architectural scheme for some simple autonomous robotic system. We are
not going to start with a concierge android right away. Our current idea
is to develop some sort of robotic vacuum cleaner prototype that does not even clean
but can bump into humans or pets and receive emotional feedback, for
example because it is too noisy or because of incorrectly prioritized work tasks.
To simplify human feedback mechanism some remote control device could be used
for example with two buttons: for positive and negative feedback. Later we can
employ some audio-analysis system to infer emotional reaction from verbal commands
to robo-cleaner. Main research question here is to what extent these emotional
mechanisms can affect robot's behavior? And we hope to establish a foundation
for desired software framework.

\bibliographystyle{splncs03}

\bibliography{neucogar}

\end{document}